 \documentclass[pmlr,twocolumn,10pt]{template/jmlr} 

\usepackage{booktabs}
\usepackage{siunitx}

\usepackage[switch]{lineno}

\usepackage{multirow}
\usepackage{multicol}
\usepackage{colortbl}

\usepackage[acronym]{glossaries}
\newacronym{cnn}{CNN}{convolution neural network}
\newacronym{resnet}{ResNet}{residual neural network}
\newacronym{dino}{DINO}{self-distillation with no labels}
\newacronym{vit}{ViT}{vision transformer}
\newacronym{vit-t}{ViT-T}{ViT-Tiny}
\newacronym{ssl}{SSL}{self-supervised learning}
\newacronym{simclr}{SimCLR}{simple framework for contrastive learning of visual representations}
\newacronym{byol}{BYOL}{bootstrap your own latent}
\newacronym{ibot}{iBOT}{image BERT pre-training with online tokenizer}
\newacronym{knn}{kNN}{$k$-nearest neighbor}

\newcommand{\equal}[1]{{\hypersetup{linkcolor=black}\thanks{#1}}}

\theorembodyfont{\upshape}
\theoremheaderfont{\scshape}
\theorempostheader{:}
\theoremsep{\newline}

\jmlrvolume{LEAVE UNSET}
\jmlryear{2024}
\jmlrsubmitted{LEAVE UNSET}
\jmlrpublished{LEAVE UNSET}
\jmlrworkshop{Machine Learning for Health (ML4H) 2024} 

\title[Towards Scalable Foundation Models for Digital Dermatology]{Towards Scalable Foundation Models for Digital Dermatology}

\author{%
\Name{Fabian Gr\"oger}$^{1,2}$ \Email{fabian.groeger@unibas.ch} \\
\Name{Philippe Gottfrois}$^1$ \Email{philippe.gottfrois@unibas.ch} \\
\Name{Ludovic Amruthalingam}$^2$ \Email{ludovic.amruthalingam@hslu.ch} \\
\Name{Alvaro Gonzalez-Jimenez}$^1$ \Email{alvaro.gonzalezjimenez@unibas.ch} \\
\Name{Simone Lionetti}$^2$ \Email{simone.lionetti@hslu.ch} \\
\Name{Luis R. Soenksen-Martinez}$^{4,5}$ \Email{soenksen@mit.edu} \\
\Name{Alexander A.\ Navarini}\equal{Joint last authorship.}$^{1,3}$ \Email{alexander.navarini@usb.ch} \\
\Name{Marc Pouly}\footnotemark[1]$^2$ \Email{marc.pouly@hslu.ch} \\
\addr 
$^1$Department of Biomedical Engineering, University of Basel, Switzerland \\
$^2$Department of Computer Science, Lucerne University of Applied Sciences and Arts, Switzerland \\
$^3$Department of Dermatology, University Hospital of Basel, Switzerland \\
$^4$J-Clinc for AI and Health, Massachusetts Institute of Technology, USA\\
$^5$Center for Bioengineering Innovation and Design, Johns Hopkins University, USA
}

\begin{document}

\frenchspacing
\maketitle

\begin{abstract}
The growing demand for accurate and equitable AI models in digital dermatology faces a significant challenge: the lack of diverse, high-quality labeled data. 
In this work, we investigate the potential of domain-specific foundation models for dermatology in addressing this challenge.
We utilize self-supervised learning (SSL) techniques to pre-train models on a dataset of over 240,000 dermatological images from public and private collections.
Our study considers several SSL methods and compares the resulting foundation models against domain-agnostic models like those pre-trained on ImageNet and state-of-the-art models such as MONET across 12 downstream tasks.
Unlike previous research, we emphasize the development of smaller models that are more suitable for resource-limited clinical settings, facilitating easier adaptation to a broad range of use cases.
Results show that models pre-trained in this work not only outperform general-purpose models but also approach the performance of models 50 times larger on clinically relevant diagnostic tasks.
To promote further research in this direction, we publicly release both the training code and the foundation models, which can benefit clinicians in dermatological applications.
\end{abstract}


\newpage
\paragraph*{Data and Code Availability.}
Datasets used for evaluation are publicly accessible and detailed in Appendix~\ref{app:Downstream-Tasks}.
Code and models to reproduce the results are available\footnote{\url{https://github.com/Digital-Dermatology/self-supervised-dermatology}}.
Together this ensures full reproducibility of the evaluation results.

\paragraph*{Institutional Review Board (IRB).}
IRB approval is not required for our research, as it utilizes existing datasets and involves no direct interaction with human subjects beyond the use of previously collected human annotations.

\section{Introduction}
\label{sec:intro}
The scarcity of high-quality, large-scale annotated data remains a significant challenge in the medical field \citep{arora2023value}.
This is due to the high costs of expert annotations~\citep{castro_causality_2020}, difficulties in reaching consensus among experts~\citep{jacob2021disentangling}, data imbalance due to rare conditions, biases in data collection~\citep{groh_evaluating_2021}, and legal constraints on the annotation process.
Transfer learning from models pre-trained on natural images such as ImageNet has become standard practice to address data scarcity \citep{baykal_transfer_2020}. 
Recently, foundation models---deep learning models trained on vast amounts of unlabeled data---have gained interest for their adaptability across a wide range of tasks \citep{awais_foundational_2023}.
Domain-specific foundation models have received particular attention in the medical field since they tend to outperform their domain-agnostic counterparts \citep{azad2023foundational}. 
Smaller foundation models are also valuable for their ease of deployment and accessibility, enabling advanced tools to be used even in resource-constrained settings~\citep{touvron2023llama}.
Although some domains such as histopathology \citep{lu2024visual}, radiology \citep{wu2023towards}, or ophthalmology \citep{zhou2023foundation} have seen great progress in this direction, dermatology is lagging behind~\citep{azad2023foundational}. 
Given the visual diversity of skin conditions and the need for equitable representation across skin tones, tailored foundation models could significantly improve both diagnostic accuracy and fairness in digital dermatology.

This paper presents preliminary results from the development of image-based foundation models for dermatology that are optimized for both efficiency and widespread accessibility. 
We focus on identifying the pre-training methodology that yields the best-performing model across 12 clinically relevant downstream tasks evaluated in terms of their frozen and low-data performance. 
Additionally, we release the models for the community to use, including training and evaluation scripts.

\section{Related Work}
The progress of foundation models is based on two key advancements: the availability of large-scale datasets and the development of learning techniques capable of training models without relying on labeled data, such as with \gls*{ssl}. 
This technique has gained popularity in the medical field because large unlabeled datasets are generally more accessible than their annotated counterparts \citep{azad2023foundational}.
The idea of \gls*{ssl} is to learn data representations using an artificial supervised objective, often called the \emph{pretext task}, created from a pool of unlabeled data. 
Representations are then generally assessed through \emph{downstream tasks} and can be used to solve relevant tasks, where annotated data are available but potentially scarce. 
In recent years, \gls*{ssl} has been widely used to learn meaningful visual foundation models without relying on manual annotations across medical disciplines~\citep{lu2024visual,wu2023towards,zhou2023foundation}. 
For a comprehensive review of \gls*{ssl} we refer to \cite{liu_self-supervised_2021} and to \cite{shurrab2022self} focusing specifically on medical applications. 
The following section introduces the \gls*{ssl} methods used in this work.

\looseness=-1
Many \gls*{ssl} methods focus on discriminative approaches, where each image is considered a separate class, and the objective is to tell them apart. 
\acrshort*{simclr}~\citep{chen_simple_2020} is a prominent example, which uses a contrastive loss to compare different views of the same image against other randomly sampled images. 
A caveat of this approach is that it requires comparing features from numerous images simultaneously, which demands large batch sizes. 
\acrshort*{byol}~\citep{grill_bootstrap_2020} is a popular method circumventing this issue with a carefully implemented asymmetric student-teacher architecture. 
Similarly, \acrshort*{dino}~\citep{caron2021emerging} compares probability-like outputs from different patches of the same image using a teacher and a student network, prominently leveraging a \gls*{vit}~\citep{dosovitskiy_image_2020} and a multi-crop augmentation strategy. 
\acrshort*{ibot}~\citep{zhou2021ibot} further builds on these ideas and explicitly exploits inherent properties of \glspl*{vit} to capture both local and global information.
Other approaches use paired multi-modal data for pre-training.
For instance, CLIP~\citep{radford2021learning} aligns visual and textual representations with a contrastive objective, jointly training image and text encoders that can be independently utilized post-training.

\section{Methodology}

\paragraph{Pre-training Data.} \label{subsub:Train-Data}
We curated a collection of unlabeled dermatology pictures from public and private sources, totaling 242,039 images. 
The collection includes modalities typically used in digital dermatology, namely dermoscopy and clinical images. 
The datasets used for pre-training are listed in Appendix~\ref{app:PretrainingData}, and include MED-NODE \citep{giotis2015a}, PH$^2$ \citep{mendoncca2013ph}, SD-260 \citep{10.1007/978-3-319-46466-4_13}, ISIC \citep{isic}, and a private collection from the University Hospital of Basel.

\paragraph{Downstream Tasks.}
We use 9 datasets for evaluation:
MED-NODE~\citep{giotis2015a}, 
PH$^2$~\citep{mendoncca2013ph},
DDI~\citep{daneshjou_disparities_2022},
Derm7pt~\citep{Kawahara2018-7pt}, 
PAD-UFES-20~\citep{pacheco_pad-ufes-20_2020}, 
SD-128~\citep{leibe_benchmark_2016}, 
PASSION~\citep{gottfrois2024PASSION}, 
HAM10000~\citep{tschandl_ham10000_2018}, 
and Fitzpatrick17k~\citep{groh_evaluating_2021}.
This results in a total of 12 diagnostic downstream tasks.
Some of the datasets purposely overlap with the pre-training data, whereas others do not, to compare generalization beyond the seen collection.
The downstream tasks were selected to represent common challenges in dermatology, such as different image modalities (clinical and dermoscopy images), different Fitzpatrick skin types, and small sample counts. 
More details on the downstream tasks can be found in Appendix~\ref{app:Downstream-Tasks}.

Final evaluations are performed on the test set provided by the dataset authors when available.
When no designated test set is provided, we set apart 15\% of the original data for testing. 
For datasets without a predefined validation set, we also split the data not used for testing into 85\%/15\% for training and validation.
Splits are performed at patient level whenever possible. 
Additionally, we remove quality issues identified by SelfClean~\citep{groger2023towards,groger_selfclean_2024} to enhance the reliability of the performance estimates.

\paragraph{Implementation.}
\looseness=-1
The foundation models in this work can be divided into two groups based on their encoder architecture, which is either a \acrshort*{cnn} or a \gls*{vit}. 
To promote comparability between the groups, we selected architectures with similar performance on ImageNet. 
For the \acrshort*{cnn}-based models, we use ResNet-50~\citep{he2016deep}, while for the \acrshort*{vit}-based models, we choose \gls*{vit-t}~\citep{dosovitskiy_image_2020} with a $16 \times 16$ patch size. 
ResNet-50 has approximately 23 million trainable parameters, and \mbox{\acrshort*{vit-t}} has 5 million.
Model configurations, including method-specific parameters and augmentation techniques, were initially taken from the original works.
We then ensured that all models converged to suitable solutions by manually tuning the optimizer, learning rate, and method-specific hyperparameters. 
Final configurations can be found in table~\ref{tab:Hyperparameters}.
Model inputs are resized to $224\times224$ pixels and normalized using the mean and standard deviation of ImageNet~\citep{deng2009imagenet}.
All models were trained until the validation loss did not improve consecutively over twenty epochs. 

The implementation is based on PyTorch~\citep{paszke2019pytorch} and the official repositories of the selected \gls*{ssl} methods.
Experiments are performed on a Nvidia DGX station, which features eight V100 GPUs, each with 32 GB of memory, 512 GB of system memory, and a CPU with 40 cores. 

\paragraph{Training and Evaluation Protocols.}
All models are evaluated on a suite of downstream tasks, adhering to standard practice in \gls*{ssl}~\citep{xu2024framework}.
Specifically, performance is computed with frozen features using both linear and \gls*{knn} models. 
Frozen evaluation examines how well representations separate a dataset attribute, indicating whether inherent properties of the downstream task were learned during pre-training, relying solely on visual information.
To further assess the models' effectiveness in low-data scenarios, we train linear and \acrshort*{knn} classifiers on randomly selected subsets of labeled data with increasing size and evaluate them all on the same test set.

We compare the domain-specific foundation models against domain-agnostic models pre-trained on ImageNet with supervision, specifically a \mbox{ResNet-50} and \gls*{vit-t}. 
Additionally, to estimate state-of-the-art performance for downstream tasks, regardless of model size or additional information used, we compare against MONET~\citep{kim2024transparent}, an image-text foundation model trained on 105,550 dermatological images paired with natural language descriptions from medical literature.
MONET is pre-trained using CLIP and uses a ViT-L image encoder with a $14\times14$ patch size, containing 304 million parameters, significantly larger than the other models compared in this work.
Moreover, it requires significantly more FLOPs than the considered models in this paper ($61.6 \times 10^9$ vs. $1.3 \times 10^9$) and has slower inference (277 vs. 3796 im/s) \citep{touvron2022three}.

\begin{figure}[t]
    \centering
    \includegraphics[width=\linewidth]{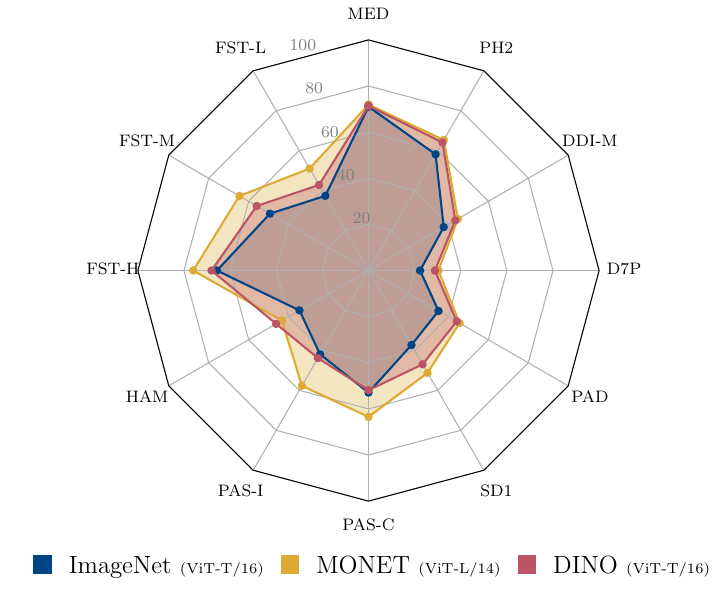}
    \caption{
    Macro-averaged test F1 for frozen \gls*{knn} evaluation of the best-performing foundation model, domain-agnostic ImageNet model, and MONET for dermatological diagnostic tasks. 
    Consult Appendix~\ref{app:Downstream-Tasks} for task details and \ref{app:Detailed-Results} for more detailed results.
    }
    \label{fig:Frozen-Performance}
\end{figure}

\begin{figure*}[t]
    \centering
    \includegraphics[width=\textwidth]{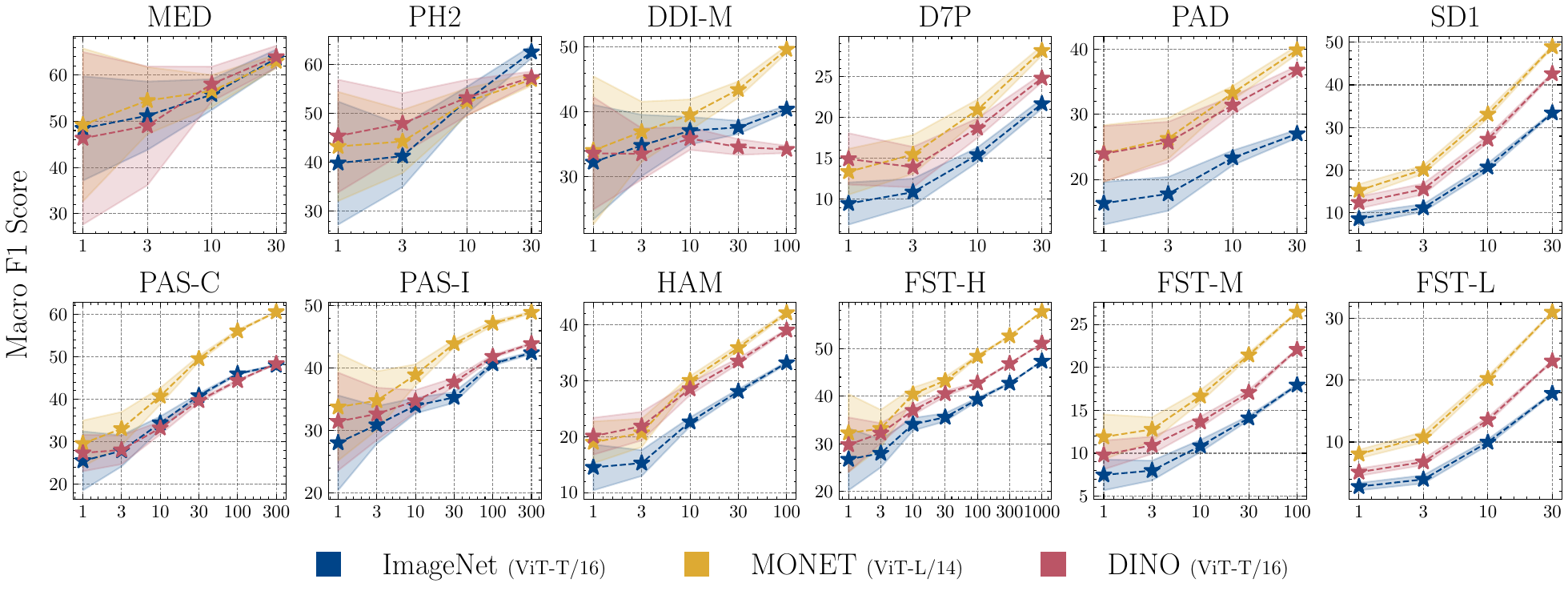}
    \caption{
    Results of a \gls*{knn} classifier on pre-trained representations when varying the number of training samples per class available for downstream tasks.
    Performance is obtained by repeating the sampling process 50 times and reporting the average and standard errors.
    }
    \label{fig:kNN-Few-Shot}
\end{figure*}

\section{Results} \label{sec:Results}
In figure~\ref{fig:Frozen-Performance}, we compare foundation models pre-trained using diverse methods on a suite of downstream tasks in terms of frozen \acrshort*{knn} performance. 
Specifically, we compare the best-performing foundation model trained in this paper, here a DINO pre-trained \gls*{vit-t}, with the best-performing domain-agnostic model, here a \gls*{vit-t} supervised pre-trained on ImageNet, and an estimation of the current state-of-the-art through MONET's performance. 
Results are obtained by repeating the evaluation with five random seeds and reporting average performance.
Detailed results for \gls*{knn} and linear evaluation can be found in table~\ref{tab:Results-Evaluation} of the appendix.
Here we explicitly focus on \gls*{knn} performance since this correlates better with fine-tuned performance and is more robust to hyperparameter choices~\citep{caron2021emerging}.

The DINO foundation model performs better in most tasks compared to the domain-agnostic ImageNet model, and is on par with state-of-the-art performance for half of the tasks.
The performance gap between DINO and state-of-the-art is most noticeable for PAS-I and PAS-C, which are very difficult.
Indeed, these only contain non-standardized, pigment-rich patients, of which the models pre-trained in this paper have seen limited amounts.
However, for other tasks that also did not overlap with the pre-training dataset, the model is generalizing well, including for those featuring pigment-rich skin, such as \mbox{DDI-M}, \mbox{FST-H}, \mbox{FST-M}, and \mbox{FST-L}.
When investigating different \gls*{ssl} strategies in table~\ref{tab:Results-Evaluation}, we find that most, except for iBOT, fall short of similar performance.
In conclusion, we observe that even models significantly smaller than MONET can achieve state-of-the-art performance with appropriate pre-training, such as DINO or iBOT.

In figure~\ref{fig:kNN-Few-Shot}, we compare different foundation models by adding a \gls*{knn} classifier on frozen features and varying the training dataset size for the downstream tasks. 
The same experiment with a linear classifier can be found in figure~\ref{fig:lin-Few-Shot} of the appendix. 
Results show that across the majority of tasks, the domain-specific DINO model performs better than the supervised domain-agnostic ImageNet model.
Additionally, for some tasks such as D7P, PAD, and HAM, the domain-specific model matches state-of-the-art performance.
In figure~\ref{fig:Utility}, we compute the utility \citep{newell2020useful} of the DINO and MONET model compared to the supervised pre-trained ImageNet model, measuring the saving in labels by the respective representations.
Specifically, utility is the ratio of additional labels needed for the supervised pre-trained model to match the performance of the others. 
For most tasks, there is a significant benefit in using the representations from the DINO model compared to that of the domain-agnostic supervised one. 
On average across all downstream tasks, the DINO model reduces the necessary annotated samples by a factor of 2.

\section{Conclusion}
In this work, we have taken steps toward developing purely visual domain-specific foundation models for digital dermatology for scalable inference. 
Through a series of experiments across diverse diagnostic tasks, we demonstrated that dermatology-specific pre-training, particularly with methods such as DINO or iBOT, can outperform and are more label-efficient than models pre-trained on generic datasets like ImageNet.
We specifically focused on smaller models and showed that they can achieve performance on par with or close to 50 times larger state-of-the-art models like MONET when subject to careful pre-training.
These smaller models are more suitable for resource-limited clinical or teledermatology settings than the current state-of-the-art models, easing the adaptability for a broad range of clinically relevant use cases.
While the models presented here show strong potential, further refinement in pre-training methodologies and expanding pre-training data to include more diverse populations and conditions will be essential for continued improvement. 
Additionally, exploring the trade-offs between model size and diagnostic performance would be valuable to further optimize the models' efficiency.
By releasing the models and tools publicly, we hope to accelerate progress in the field of AI for dermatology, as these models can serve as a starting point for future applications and lower the need for annotated samples.


\bibliography{groeger24}

\appendix
\section{Pre-training Dataset}
\label{app:PretrainingData}

This section details the datasets used for pre-training.
Public datasets without a license are under the public domain mark.

\begin{itemize}
\item 
    \textbf{MED-NODE}~\citep{giotis2015a} 
    features 170 clinical images for skin cancer detection by the University Medical Center Groningen, Netherlands (CC BY 4.0).

\item 
    \textbf{PH$^2$ Database}~\citep{mendoncca2013ph} 
    features 200 dermoscopy images for melanocytic lesion classification by the Hospital Pedro Hispano in Matosinhos, Portugal.

\item 
    \textbf{Derm7pt}~\citep{Kawahara2018-7pt} 
    features 2,022 dermoscopy and clinical images for skin condition diagnosis (CC BY-NC-SA 4.0).
    
\item 
    \textbf{SD-260}~\citep{10.1007/978-3-319-46466-4_13} 
    features 12,583 clinical images for skin condition diagnosis collected from DermQuest.

\item 
    \textbf{ISIC}~\citep{isic} 
    features 107,208 dermoscopy images of pigmented skin lesions and features almost only low-pigmented skin (CC-BY-NC).
    
\item 
    A private collection of 119,858 clinical images reflecting the data distribution encountered in Swiss hospitals. 
    Pictures were taken using diverse reflex cameras by trained photographers, anonymized, and used with approval EKNZ-2018-01074 from an ethical committee according to Swiss regulations.
\end{itemize}

\section{Downstream Tasks}
\label{app:Downstream-Tasks}
This section lists the suite of datasets used for evaluation.
Bold indicates the respective downstream tasks.

\begin{itemize}
\item 
    \textbf{MED-NODE (MED)}~\citep{giotis2015a} 
    features 170 clinical images for skin cancer detection by the University Medical Center Groningen, Netherlands (CC BY 4.0). The images are categorized into melanoma and naevus.

\item 
    \textbf{PH$^2$ Database (PH2)}~\citep{mendoncca2013ph} 
    features 200 dermoscopy images for melanocytic lesion classification by the Hospital Pedro Hispano in Matosinhos, Portugal. The images are categorized into common nevi, atypical nevi, and melanomas.

\item
    \textbf{DDI}~\citep{daneshjou_disparities_2022} features 656 clinical images for skin cancer detection and rare disease classification by the Stanford Clinics (Stanford's University dataset research agreement). 
    The images are categorized into benign or malignant lesions \mbox{(\textbf{DDI-M})}.

\item 
    \textbf{Derm7pt (D7P)}~\citep{Kawahara2018-7pt} 
    features 2,022 dermoscopy and clinical images for skin condition diagnosis (CC BY-NC-SA 4.0). 
    The images are categorized into 16 diagnoses.
    
\item 
    \textbf{PAD-UFES-20 (PAD)} \citep{pacheco_pad-ufes-20_2020} 
    features clinical images captured by smartphones (CC BY 4.0 license). 
    The dataset consists of 1,373 patients, 1,641 skin lesions, and 2,298 images for six disease diagnoses.

\item 
    \textbf{PASSION}~\citep{gottfrois2024PASSION} features 4,901 clinical images of 1,653 pigment-rich patients from Sub-Saharan countries (CC BY-NC 4.0 license).
    The dataset is categorized into the four common pediatric conditions (\mbox{\textbf{PAS-C}}) and impetigo cases (\mbox{\textbf{PAS-I}}).

\item
    \textbf{SD-128 (SD1)}~\citep{leibe_benchmark_2016} features 5,619 clinical images for skin condition diagnosis collected from DermQuest. 
    The images are categorized into 123 diseases.
    
\item 
    \textbf{HAM10000 (HAM)} \citep{tschandl_ham10000_2018} 
    features of 10,015 dermatoscopic images collected from different populations and institutions for seven diagnostic categories of pigmented lesions (CC BY-NC 4.0). 
    
\item 
    \textbf{Fitzpatrick17k} \citep{groh_evaluating_2021} 
    features 16,577 clinical images (CC BY-NC-SA 3.0). 
    The dataset's conditions are separated into three different granularity levels: high (\textbf{FST-H}), mid (\textbf{FST-M}), and low (\textbf{FST-L}).
\end{itemize}

\section{Hyperparameters}
\label{app:Hyperparameters}
Table~\ref{tab:Hyperparameters} details the hyperparameters for the pre-training methods used to train the respective foundation models.

\begin{table*}
    \centering
    \caption{
    Hyperparameters used for each \gls*{ssl} method, where ``n.a.'' indicates that this parameter is not used.
    }
    \label{tab:Hyperparameters}
    \begin{tabular}{l rrrr}
        \toprule
        \textbf{Hyperparameter} 
            & \textbf{SimCLR} 
            & \textbf{BYOL} 
            & \textbf{DINO} 
            & \textbf{iBOT}\\
        \midrule
        Encoder & ResNet-50 & ResNet-50 & \acrshort*{vit-t} & \acrshort*{vit-t} \\
        Batch size & 640 & 60 & 260 & 224\\
        Epochs & 100 & 100 & 100 & 100 \\
        Optim & Adam & Adam & AdamW & AdamW \\
        Learning rate (lr) & 0.00003 & 0.003 & 0.0005 & 0.005 \\
        
        Min. lr. & 1e-6 & 1e-6 & 1e-6 & 5e-4 \\
        Warmup epochs lr. & 10 & 10 & 10 & 10 \\
        Weight decay (wd) & 1e-6 & n.a. & 0.04 & 0.04 \\
        Weight decay end & 0.4 & n.a. & 0.4 & 0.4 \\
        
        Use lr. scheduler & True & True & True & True  \\
        Use wd. scheduler & True & False & True & True \\
        
        Momentum teacher & n.a. & 0.996 & 0.9995 & 0.996 \\

        Global crops scale & n.a. & n.a. & (0.7, 1.) & (0.7, 1.) \\
        Local crops scale & n.a. & n.a. & (0.05, 0.4) & (0.05, 0.4) \\
        N.o. global, local crops & n.a. & n.a. & (2, 12) & (2, 12) \\
        \bottomrule
    \end{tabular}
\end{table*}

\section{Detailed Results} \label{app:Detailed-Results}
\begin{figure*}[t]
    \centering
    \includegraphics[width=\textwidth]{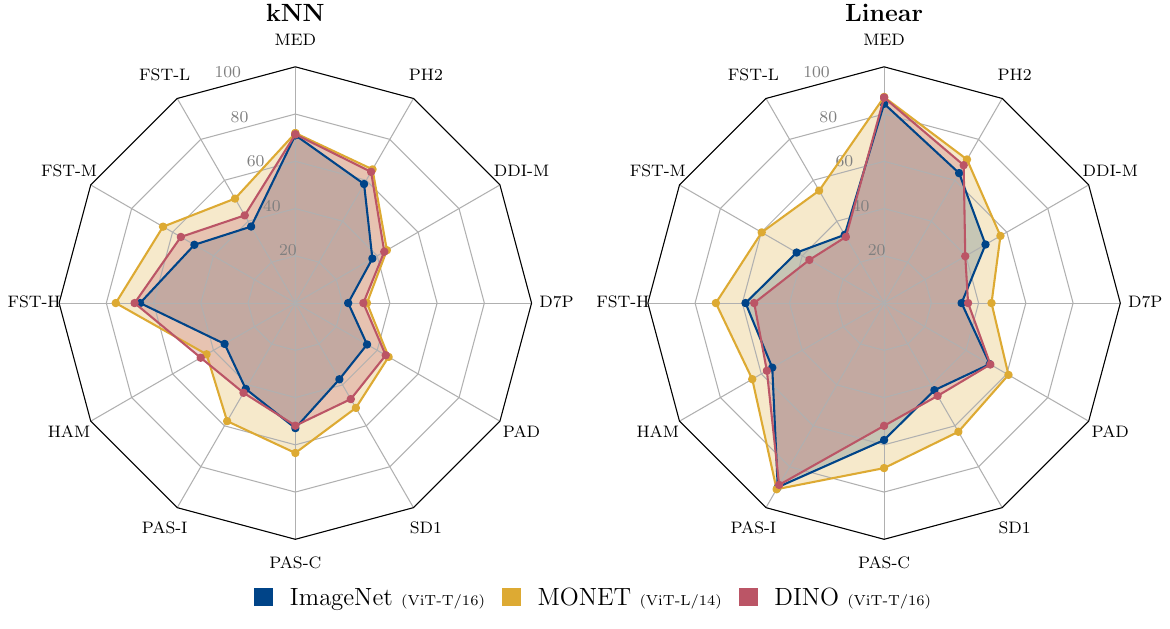}
    \caption{
    Macro-averaged test F1 for frozen \gls*{knn} and linear evaluation of the best-performing foundation model, domain-agnostic ImageNet model, and MONET for dermatological diagnostic tasks. 
    Consult Appendix~\ref{app:Downstream-Tasks} for task details and \ref{app:Detailed-Results} for more detailed results.
    }
    \label{fig:SpiderPlot}
\end{figure*}

Table~\ref{tab:Results-Evaluation} details the results from the frozen evaluation on the suite of downstream tasks.
Consult Appendix~\ref{app:Downstream-Tasks} for details on downstream tasks and section \ref{sec:Results} for the discussion on the results.
Additionally, figure~\ref{fig:SpiderPlot} visualizes both \gls*{knn} and linear performance, similarly as done for figure~\ref{fig:Frozen-Performance}.

\begin{figure*}[t]
    \centering
    \includegraphics[width=\textwidth]{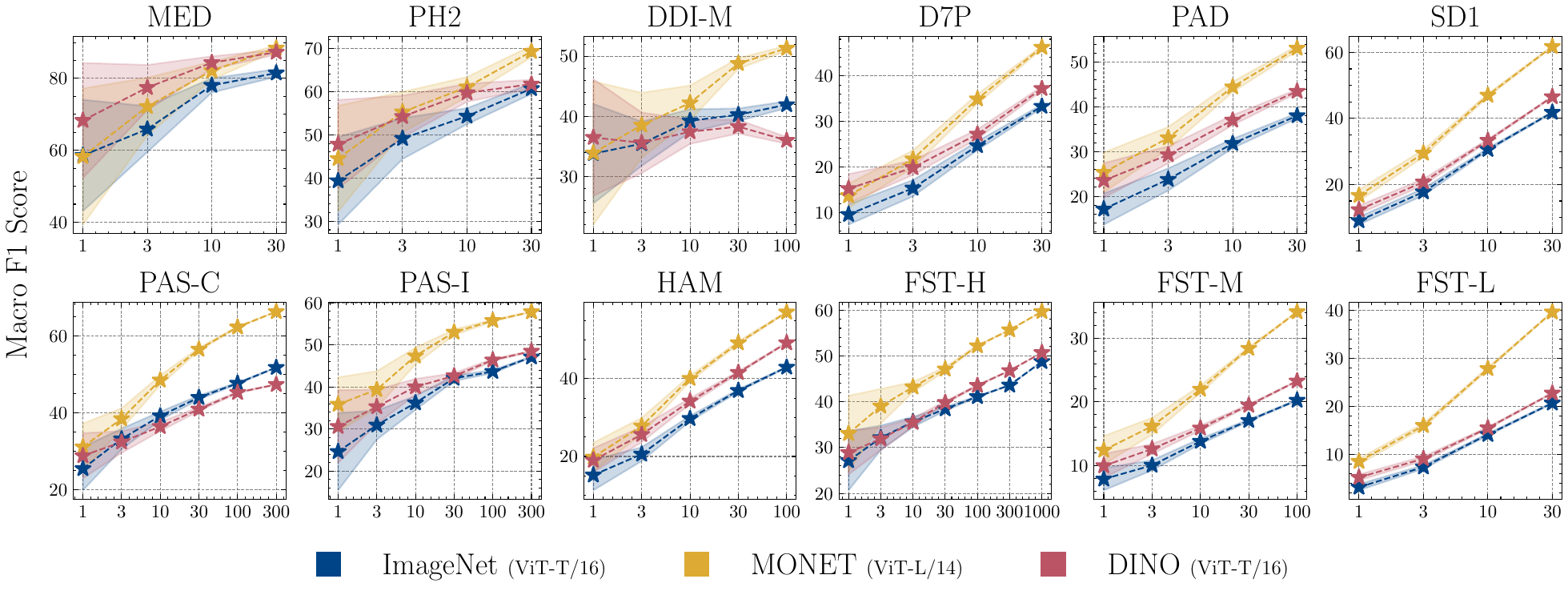}
    \caption{
    Results of a linear classifier on pre-trained representations when varying the number of training samples per class available for downstream tasks.
    Performance is obtained by repeating the sampling process 50 times and reporting the average and standard errors.
    }
    \label{fig:lin-Few-Shot}
\end{figure*}

Figure~\ref{fig:lin-Few-Shot} shows the evaluation in low-data scenarios with linear classifiers, similar to figure~\ref{fig:kNN-Few-Shot}.
Figure~\ref{fig:Utility} visualizes the utility of the representations for both \gls*{knn} and linear evaluation in low-data scenarios as discussed in section \ref{sec:Results}.

\begin{table*}
    \caption{
    Macro-averaged test F1 scores of pre-trained models for twelve dermatology diagnostic tasks.
    Bold values correspond to the best scores.
    ANOVA was used to compare the mean F1 scores across multiple models, followed by Tukey's post-hoc test to identify significant pairwise differences \mbox{($p<0.05$)} denoted with a $^*$.
    }
    \label{tab:Results-Evaluation}

    \centering

    \resizebox{\linewidth}{!}{%
    \begin{tabular}{lll cc cc cc cc}%
        \toprule
        &&
        & \multicolumn{2}{c}{\bfseries MED} 
        & \multicolumn{2}{c}{\bfseries PH2}
        & \multicolumn{2}{c}{\bfseries DDI-M} 
        & \multicolumn{2}{c}{\bfseries D7P} \\
        
        \cmidrule(lr){4-5}
        \cmidrule(lr){6-7}
        \cmidrule(lr){8-9}
        \cmidrule(lr){10-11}
        
        \bfseries Pre-training 
        & \bfseries Dataset
        & \bfseries Architecture
        & lin. & kNN
        & lin. & kNN
        & lin. & kNN
        & lin. & kNN \\
        \midrule
        Supervised & ImageNet & ResNet-50
            & $84.5 \pm 2.9$ & $64.6 \pm 2.1$
            & $61.0 \pm 4.0$ & $47.7 \pm 7.1$
            & $50.3 \pm 3.2$ & $39.1 \pm 0.9$
            & $32.6 \pm 1.0$ & $19.9 \pm 1.7$ \\
        Supervised & ImageNet & \acrshort*{vit-t}/16
            & $84.4 \pm 0.9$ & $71.1 \pm 3.9$
            & $63.5 \pm 2.8$ & $58.2 \pm 5.7$
            & $49.6 \pm 3.2$ & $37.7 \pm 2.7$
            & $32.8 \pm 1.5$ & $22.4 \pm 1.8$ \\
        MONET & Derma Lit. & ViT-L/14
            & $87.2 \pm 2.6$ & $71.9 \pm 2.8$
            & $\mathbf{70.2 \pm 1.3}$ & $\mathbf{65.3 \pm 4.9}$
            & $\mathbf{56.9 \pm 5.7}$ & $\mathbf{44.7 \pm 1.9}$
            & $\mathbf{45.5 \pm 2.3^*}$ & $\mathbf{30.3 \pm 1.2}$ \\ 
        \midrule
        SimCLR & Derma Img. & ResNet-50
            & $75.0 \pm 0.0$ & $71.4 \pm 4.5$
            & $32.9 \pm 1.9$ & $50.8 \pm 3.1$
            & $0.0 \pm 0.0$ & $31.2 \pm 1.6$
            & $3.9 \pm 0.1$ & $15.3 \pm 2.1$ \\
        BYOL & Derma Img. & ResNet-50
            & $77.4 \pm 3.1$ & $67.5 \pm 3.8$
            & $35.0 \pm 3.0$ & $60.2 \pm 6.9$
            & $0.0 \pm 0.0$ & $29.3 \pm 3.6$
            & $7.2 \pm 0.1$ & $21.7 \pm 0.8$ \\ 
        DINO & Derma Img. & \acrshort*{vit-t}/16
            & $87.0 \pm 2.3$ & $71.4 \pm 2.2$
            & $67.4 \pm 3.3$ & $64.1 \pm 1.8$
            & $39.7 \pm 3.0$ & $43.5 \pm 4.3$
            & $35.6 \pm 0.9$ & $28.8 \pm 1.8$ \\
        iBOT & Derma Img. & \acrshort*{vit-t}/16
            & $\mathbf{88.7 \pm 2.0}$ & $\mathbf{77.5 \pm 4.3}$
            & $61.7 \pm 2.3$ & $61.8 \pm 4.0$
            & $42.4 \pm 5.0$ & $38.2 \pm 5.5$
            & $31.3 \pm 1.7$ & $26.1 \pm 1.1$ \\ 
            
        \midrule

        &&
        & \multicolumn{2}{c}{\bfseries PAD} 
        & \multicolumn{2}{c}{\bfseries PAS-C} 
        & \multicolumn{2}{c}{\bfseries PAS-I} 
        & \multicolumn{2}{c}{\bfseries SD1} \\
        
        \cmidrule(lr){4-5}
        \cmidrule(lr){6-7}
        \cmidrule(lr){8-9}
        \cmidrule(lr){10-11}
        
        \bfseries Pre-training 
        & \bfseries Dataset
        & \bfseries Architecture
        & lin. & kNN
        & lin. & kNN
        & lin. & kNN
        & lin. & kNN \\
        \midrule
        Supervised & ImageNet & ResNet-50
            & $55.4 \pm 1.0$ & $40.4 \pm 1.8$
            & $56.9 \pm 0.0$ & $55.8 \pm 0.3$
            & $87.1 \pm 0.0$ & $44.5 \pm 0.8$
            & $46.2 \pm 0.3$ & $39.0 \pm 0.5$ \\
        Supervised & ImageNet & \acrshort*{vit-t}/16
            & $51.7 \pm 1.6$ & $35.1 \pm 1.4$
            & $58.0 \pm 0.0$ & $52.9 \pm 0.2$
            & $90.0 \pm 0.0$ & $42.0 \pm 0.5$
            & $42.6 \pm 0.3$ & $37.3 \pm 0.6$ \\
        MONET & Derma Lit. & ViT-L/14
            & $\mathbf{60.8 \pm 1.4}$ & $\mathbf{45.6 \pm 1.4}$
            & $\mathbf{69.9 \pm 0.0^*}$ & $\mathbf{63.5 \pm 0.1^*}$
            & $\mathbf{91.0 \pm 0.0^*}$ & $\mathbf{57.7 \pm 0.8^*}$
            & $\mathbf{62.9 \pm 0.5^*}$ & $\mathbf{51.2 \pm 0.8^*}$ \\ 
        \midrule
        SimCLR & Derma Img. & ResNet-50
            & $18.0 \pm 0.5$ & $28.6 \pm 1.7$
            & $27.6 \pm 0.0$ & $40.0 \pm 0.2$
            & $88.8 \pm 0.0$ & $29.8 \pm 0.6$
            & $0.1 \pm 0.0$ & $16.2 \pm 0.3$ \\
        BYOL & Derma Img. & ResNet-50
            & $30.1 \pm 0.9$ & $34.9 \pm 1.4$
            & $52.2 \pm 0.0$ & $54.5 \pm 0.4$
            & $88.8 \pm 0.0$ & $44.6 \pm 1.0$
            & $3.9 \pm 0.3$ & $39.0 \pm 1.1$ \\ 
        DINO & Derma Img. & \acrshort*{vit-t}/16
            & $51.9 \pm 1.7$ & $44.2 \pm 1.9$
            & $52.0 \pm 0.6$ & $51.9 \pm 0.4$
            & $88.9 \pm 0.2$ & $43.9 \pm 1.5$
            & $45.3 \pm 0.5$ & $46.9 \pm 0.4$ \\
        iBOT & Derma Img. & \acrshort*{vit-t}/16
            & $50.6 \pm 1.4$ & $39.5 \pm 0.5$
            & $49.1 \pm 0.6$ & $50.1 \pm 0.6$
            & $88.4 \pm 0.2$ & $43.8 \pm 1.1$
            & $44.9 \pm 0.8$ & $46.4 \pm 0.6$ \\ 
        \midrule

        &&
        & \multicolumn{2}{c}{\bfseries HAM} 
        & \multicolumn{2}{c}{\bfseries FST-H} 
        & \multicolumn{2}{c}{\bfseries FST-M} 
        & \multicolumn{2}{c}{\bfseries FST-L} \\
        
        \cmidrule(lr){4-5}
        \cmidrule(lr){6-7}
        \cmidrule(lr){8-9}
        \cmidrule(lr){10-11}
        
        \bfseries Pre-training 
        & \bfseries Dataset
        & \bfseries Architecture
        & lin. & kNN
        & lin. & kNN
        & lin. & kNN
        & lin. & kNN \\
        \midrule
        Supervised & ImageNet & ResNet-50
            & $57.2 \pm 0.0$ & $43.3 \pm 0.3$
            & $61.1 \pm 0.2$ & $67.2 \pm 0.4$
            & $45.0 \pm 1.1$ & $52.6 \pm 0.8$
            & $39.7 \pm 0.5$ & $40.1 \pm 0.4$ \\
        Supervised & ImageNet & \acrshort*{vit-t}/16
            & $54.7 \pm 0.0$ & $34.6 \pm 0.3$
            & $58.6 \pm 0.4$ & $65.7 \pm 0.6$
            & $42.8 \pm 0.9$ & $49.3 \pm 0.6$
            & $33.4 \pm 0.8$ & $37.4 \pm 0.6$ \\
        MONET & Derma Lit. & ViT-L/14
            & $\mathbf{64.4 \pm 0.0^*}$ & $43.4 \pm 0.2$
            & $\mathbf{71.2 \pm 0.2^*}$ & $\mathbf{76.0 \pm 0.6^*}$
            & $\mathbf{59.8 \pm 0.4^*}$ & $\mathbf{64.7 \pm 1.0^*}$
            & $\mathbf{55.0 \pm 0.4^*}$ & $\mathbf{51.1 \pm 0.2^*}$ \\ 
        \midrule
        SimCLR & Derma Img. & ResNet-50
            & $15.4 \pm 0.0$ & $28.0 \pm 0.5$
            & $30.1 \pm 0.2$ & $49.7 \pm 0.3$
            & $8.8 \pm 0.0$ & $29.3 \pm 0.8$
            & $0.3 \pm 0.0$ & $18.9 \pm 0.3$ \\
        BYOL & Derma Img. & ResNet-50
            & $28.8 \pm 0.0$ & $41.0 \pm 0.4$
            & $45.9 \pm 0.3$ & $66.1 \pm 0.4$
            & $17.3 \pm 0.2$ & $52.0 \pm 0.9$
            & $3.9 \pm 0.1$ & $39.1 \pm 0.3$ \\ 
        DINO & Derma Img. & \acrshort*{vit-t}/16
            & $57.3 \pm 0.8$ & $\mathbf{46.3 \pm 0.2^*}$
            & $55.0 \pm 0.2$ & $68.0 \pm 0.5$
            & $36.5 \pm 0.5$ & $56.0 \pm 0.2$
            & $32.3 \pm 0.6$ & $42.9 \pm 0.6$ \\
        iBOT & Derma Img. & \acrshort*{vit-t}/16
            & $57.9 \pm 0.7$ & $42.0 \pm 0.5$
            & $53.6 \pm 0.3$ & $68.5 \pm 0.6$
            & $33.8 \pm 0.9$ & $55.8 \pm 0.6$
            & $29.8 \pm 0.5$ & $42.8 \pm 0.3$ \\ 
        \bottomrule
    \end{tabular}
    }
\end{table*}

\begin{figure*}[htbp]
\floatconts
  {fig:Utility}
  {\caption{
    Utility score \citep{newell2020useful} of the \gls*{knn} and linear classifiers on pre-trained representations of MONET and DINO from figures \ref{fig:kNN-Few-Shot} and \ref{fig:lin-Few-Shot} compared against those from a supervised pre-trained ViT-T on ImageNet.
    Utility quantifies the effectiveness of the representations as the saving in labels.
    Specifically, to achieve the same accuracy with other representations, the ratio of how many more labels would be needed.
    It is positive if there is a benefit, negative if there is a drawback, zero if there is no difference in using the representations, and infinite when there is not any number of labeled samples such that the performance matches.
  }}
  {%
    \subfigure[\Acrlong*{knn} classifiers]{\label{fig:knn-Utility}%
      \includegraphics[width=\textwidth]{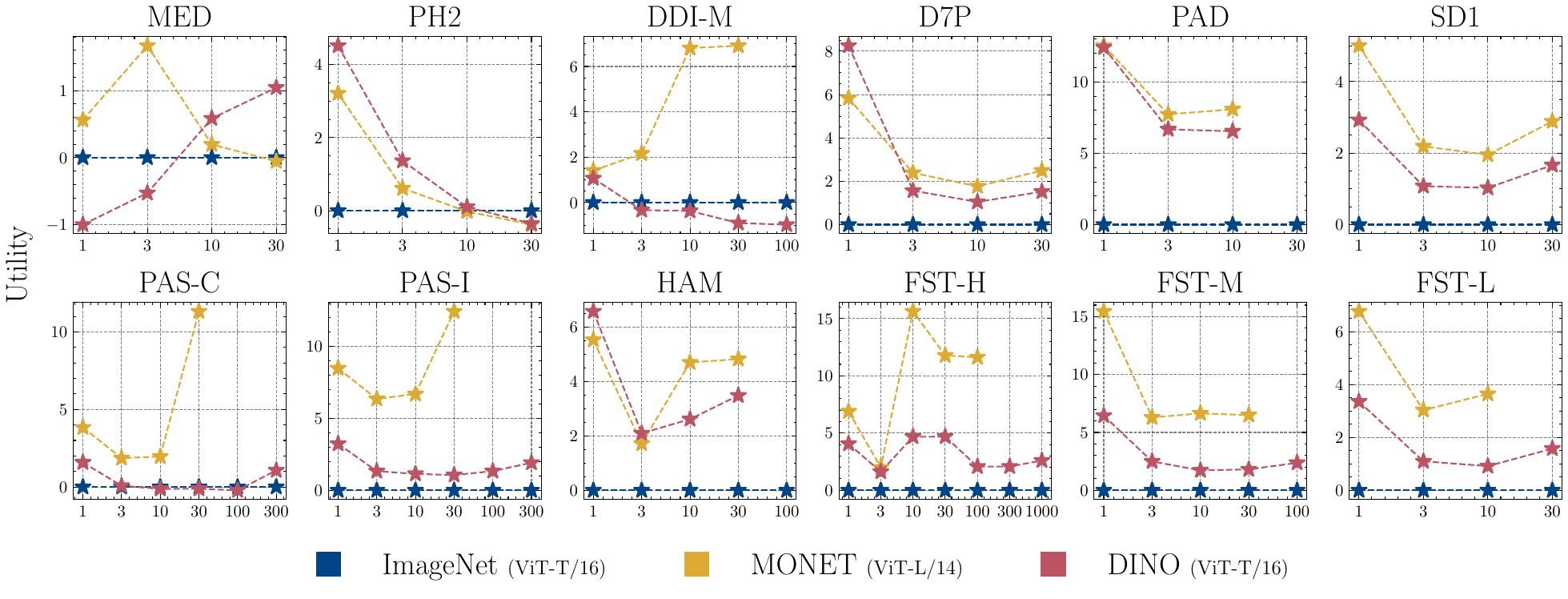}}%
      
    \subfigure[Linear classifiers]{\label{fig:lin-Utility}%
      \includegraphics[width=\textwidth]{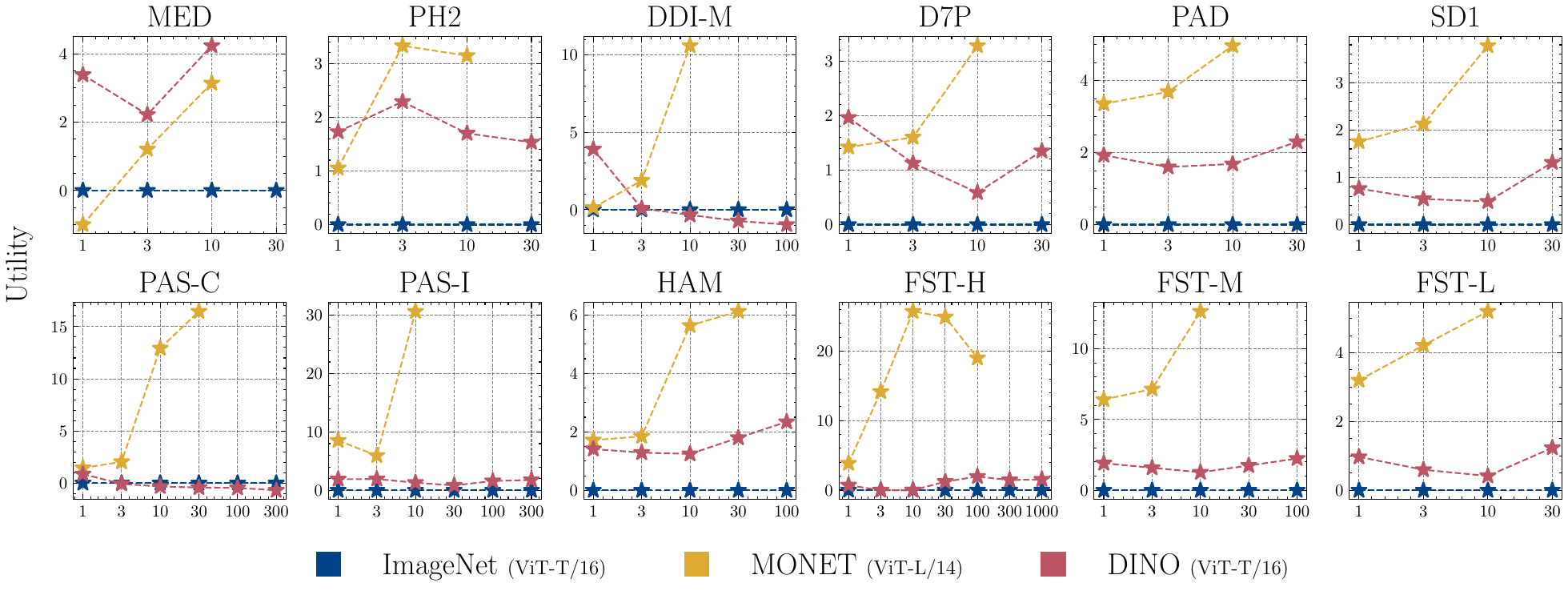}}
  }
\end{figure*}

\section{PyPi Package}

All checkpoints are released to make the foundation models widely available, and a dedicated package has been developed, enabling quick loading and instantiation of a foundation model to be used for downstream tasks.
The package can be installed using \texttt{pip} via and will be released after the decision:

\begin{verbatim}
	pip install self-supervised-dermatology
\end{verbatim}

Afterward, the package can be imported and used to obtain the pre-trained models, which can be used as PyTorch models.

\begin{verbatim}
	from self-supervised-dermatology import Embedder

	model = Embedder.load_pretrained('SimCLR')
	model = Embedder.load_pretrained('BYOL')
	model = Embedder.load_pretrained('DINO')
	model = Embedder.load_pretrained('iBOT')

	rand_img = torch.rand(1, 3, 224, 224)
	embedding = model(rand_img)
\end{verbatim}

\end{document}